\documentclass[letterpaper, 10 pt, conference]{ieeeconf}  % Comment this line out if you need a4paper

\IEEEoverridecommandlockouts                              % This command is only needed if 
\usepackage{graphicx} % for pdf, bitmapped graphics files
\usepackage{amsmath} % assumes amsmath package installed
\usepackage{amssymb}  % assumes amsmath package installed
\usepackage{xcolor} % for setting colors
\usepackage{xurl}
\usepackage{caption}
\usepackage{balance}

\title{\LARGE \bf
Single-Shot Metric Depth from Focused Plenoptic Cameras
}

\author{Blanca Lasheras-Hernandez\textsuperscript{\textdagger \S}, \, Klaus~H.\ Strobl\textsuperscript{\textdagger}, \, Sergio Izquierdo\textsuperscript{\S}, \\ Tim Bodenm\"{u}ller\textsuperscript{\textdagger}, \, Rudolph Triebel\textsuperscript{\textdagger \textdaggerdbl}, \, and Javier Civera\textsuperscript{\S}% <-this % stops a space
\thanks{\textdagger \quad Institute of Robotics and Mechatronics, German Aerospace Center (DLR) 
        {\tt\scriptsize <name>.<surname(s)>@dlr.de}}%
\thanks{\S \quad I3A, Universidad de Zaragoza 
        {\tt\scriptsize \{izquierdo,jcivera\}@unizar.es}}%
\thanks{\textdaggerdbl \quad Karlsruhe Institute of Technology 
        {\tt\scriptsize <name>.<surname>@kit.edu}}%
}

\begin{document}

% Macros for comments
\newcommand{\blanca}[1]{\textcolor{red}{[Blanca]: #1}}

\maketitle
\thispagestyle{empty}
\pagestyle{empty}

%%%%%%%%%%%%%%%%%%%%%%%%%%%%%%%%%%%%%%%%%%%%%%%%%%%%%%%%%%%%%%%%%%%%%%%%%%%%%%%%
\begin{abstract}

Metric depth estimation from visual sensors is crucial for robots to perceive, navigate, and interact with their environment. Traditional range imaging setups, such as stereo or structured light cameras, face hassles including calibration, occlusions, and hardware demands, with accuracy limited by the baseline between cameras. Single- and multi-view monocular depth offers a more compact alternative, but is constrained by the unobservability of the metric scale. Light field imaging provides a promising solution for estimating metric depth by using a unique lens configuration through a single device. %This allows depth measurement via depth-dependent refraction. 
However, its application to single-view dense metric depth is under-addressed mainly due to the technology's high cost, the lack of public benchmarks, and proprietary geometrical models and software. 

Our work explores the potential of focused plenoptic cameras for dense metric depth. 
We propose a novel pipeline that predicts metric depth from a single plenoptic camera shot by first generating a sparse metric point cloud using a neural network, which is then used to scale and align a dense relative depth map regressed by a foundation depth model, resulting in a dense metric depth. To validate it, we curated the Light Field \& Stereo Image Dataset\footnote{Dataset available at {\scriptsize \url{https://zenodo.org/records/14224205}}.} (LFS) of real-world light field images with stereo depth labels, filling a current gap in existing resources. Experimental results show that our pipeline produces accurate metric depth predictions, laying a solid groundwork for future research in this field.\footnote{ Work partially supported by the DLR Impulse Project \textit{SaiNSOR}.}

\end{abstract}

%%%%%%%%%%%%%%%%%%%%%%%%%%%%%%%%%%%%%%%%%%%%%%%%%%%%%%%%%%%%%%%%%%%%%%%%%%%%%%%%

\section{Introduction}

Within the last decades, computer vision has become a relevant domain as numerous applications demand visual perception. For example, it allows robots to navigate, understand, and interact with the environment. Particularly, accurate depth estimation becomes critical for applications where safety and reliability are paramount, such as autonomous driving~\cite{lasheras2022drivernn} and robotic manipulation~\cite{fuchs2010cooperative}.

From a single pinhole camera, we can estimate the 3D reconstruction of an unknown scene using monocular structure-from-motion (SfM), but this is subject to scale ambiguity~\cite{hartley2003multiple}. In learning-based single-view approaches, scale is also unobservable for self-supervised models~\cite{godard2019digging}, as their losses are based on the same multi-view geometric constraints. For supervised ones, the scale accuracy depends on the training and test data~\cite{li2024binsformer}. Multi-sensor setups, such as visual-inertial, stereo, or multi-camera, do observe the scale. However, in addition to their higher degree of hardware complexity, they require a precise alignment and calibration, and their accuracy is limited by the baseline between the cameras, which is conditioned by the applications. In the case of visual-inertial systems, their state may not be observable for certain motions.

Light field cameras represent a promising alternative for metric depth estimation. Their inner configuration involves a microlens array (MLA) placed between the main lens and the sensor. This allows the acquisition not only of the intensity but also the direction of light rays that interact with the sensor, by acquiring range-dependent sub-aperture images arranged in a grid pattern~\cite{feng20183d}. This unique internal structure allows the capture of multiple views and ranges of local areas with a single device and a single shot, overcoming the limitations of traditional monocular and multi-sensor setups. Additional advantages include their higher light capture (as they produce focused images even with large lens apertures), the elimination of disocclusion issues (since the scene is captured in 3D using a single main lens), and the absence of moving parts (unlike optical refocusing systems). These features have already found applications in robotics, such as on-orbit servicing and robotic exploration \cite{lingenauber:2017aero,lingenauber:2019aero}.

\begin{figure} 
    \centering
    \includegraphics[width=\columnwidth]{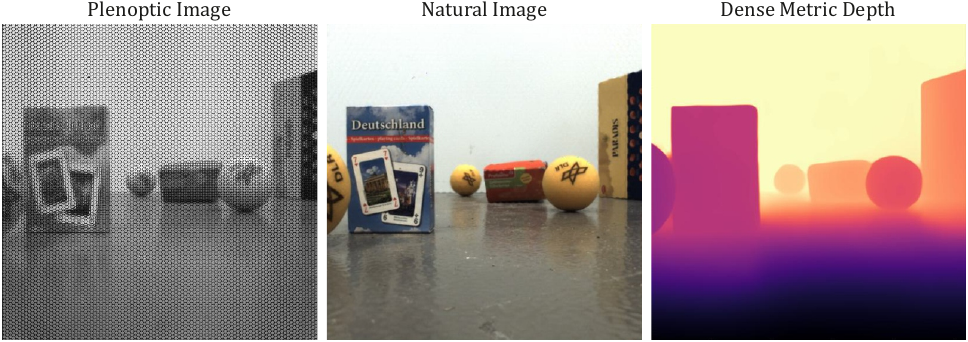}
    \caption{\textit{Left:} Plenoptic image from a light field camera, displaying the microlens pattern (see detail in Fig.~\ref{fig:flower_stack}). \textit{Center:} Corresponding natural image, synthesized from the central viewpoint of the plenoptic camera. \textit{Right:} Our single-shot metric depth map, at the true scale of the scene.}
    \label{fig:teaser}
    \vspace*{-1em}
\end{figure}

In this paper, we explore the potential of a focused plenoptic camera for single-view metric depth estimation (see Fig.~\ref{fig:teaser}). 
Specifically, we present an end-to-end pipeline that leverages light field images to resolve the scale ambiguity of a learning-based monocular depth regressor, hence producing dense metric depth maps. We believe that our approach, which leverages both, priors from a foundation model with global receptive fields and geometric cues between multiple microlenses, is bound to outperform geometric triangulation methods from local correspondences in plenoptic images. To benchmark our approach, we curated a novel real-world Light Field \& Stereo Image Dataset (LFS) that we release with this paper. Our results show that we outperform related baselines, specifically the manufacturer's software for depth, sparse and presumably based only on geometry, and the state-of-the-art foundation model Depth Anything.

\begin{figure}
    \centering
    \includegraphics[width=0.97\columnwidth]{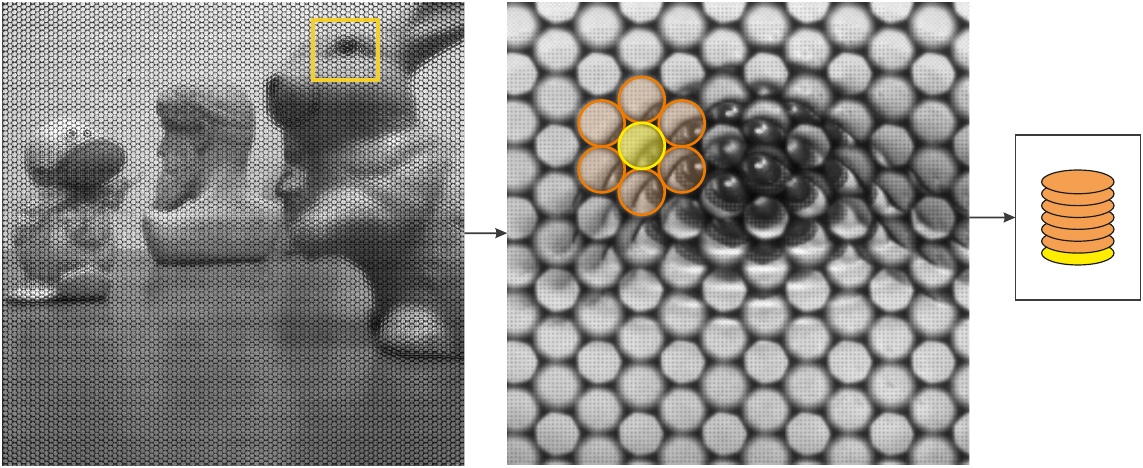}
    \caption{\textit{Flower stack} illustration. Each \textit{flower stack} is constructed by piling a central microlens and another six in the ring surrounding it. Each stacked microlens is debayered into a 3-channel RGB image.}
    \label{fig:flower_stack}
\end{figure}

\section{Related Work}

Depth estimation is a traditional topic in computer vision. \textbf{Multi-view approaches} have been extensively studied, offer improvements by capturing data from multiple vantage points rather than just two, as in stereo vision~\cite{Sinha2014}. In practice, most often the sum of squared differences is used to evaluate differences across images, facing the challenge of balancing geometric accuracy with robustness to perspective changes of local image patterns~\cite{gallup2008variable,fontan2022model}. Techniques such as Kalman filtering enhance depth estimation by processing sequential observations, while aggregation methods like sliding windows or global optimization become necessary when fewer images are available~\cite{kang2001handling}. Structure-from-motion batch pipelines like COLMAP~\cite{schonberger2016structure} further refine depth estimation by selecting views and ensuring geometric consistency between multiple depth maps, allowing for better handling of occlusions~\cite{szeliski1999multi, kang2004extracting, maitre2008symmetric, zhang2008recovering, schonberger2016pixelwise}.

\textbf{Single-view depth} is geometrically ill-posed~\cite{chiangAndBohg2022}. Various computational methods, inspired by human visual perception, have been developed to tackle this challenge using visual cues such as perspective and and visual appearance through lighting and occlusion~\cite{chen2022photoelectric, zhou2021review}. Deep neural networks have shown impressive results at predicting per-pixel depth or disparity by learning image patterns from databases with and without supervision~\cite{eigen2014depth,godard2019digging,facil2019cam,bhat2021adabins,rodriguez2023lightdepth,li2024binsformer,depthanything}.

\begin{figure}
    \centering
    \includegraphics[width=0.97\columnwidth]{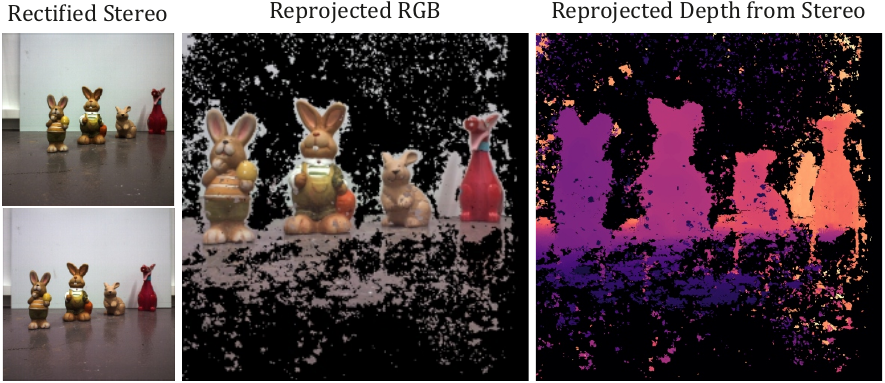}
    \caption{Visual results produced through the stereo processing pipeline to obtain suitable ground-truth depth values from stereo images, reprojected onto the plenoptic camera's calibrated reference frame.}
    \label{fig:stereoPipeline}
\end{figure}

\textbf{Depth from light field cameras}. Capturing light field images in a single exposure combines the simplicity of single-shot setups with the depth information coming from local calibrated multiple views, thereby potentially adding scale information to single-view methods. They have evolved from the plenoptic camera 1.0, which focused the main lens at the microlens array (MLA) plane, allowing for post-capture processing but at a lower resolution, to the plenoptic camera 2.0, which focuses the MLA on the main lens's image plane, enhancing image quality by balancing spatial and angular resolution. For both devices, however, a recurring challenge is the scarcity of available data. This is due to the limited availability of light field cameras and the difficulties in setting up ground-truth measuring systems for supervised learning. While camera arrays offer an effective answer, they are costly and impractical for widespread use. Additionally, plenoptic cameras are hindered by their reliance on proprietary software, high costs, and limited datasets~\cite{standofdLytro, robertson2018, rxlive}.

Some works have addressed the depth estimation task using classical methods developed for earlier light field capturing devices~\cite{ng2005light, zhang2017ray, hahne2018baseline, liu2020feature}. However, there are no available depth estimation methods for the plenoptic camera 2.0, primarily due to the lack of a publicly available geometric model and the scarcity of available real-world data, which impedes the development of learning-based approaches.
This shortage stems from the limited availability of light field capturing devices and from the difficulty of obtaining precise ground truth, as it requires an external system and data registration. Moreover, the technology incorporated into these cameras is conditioned by the dependence on the manufacturer's proprietary camera model and software for decoding raw data~\cite{rxlive}. In addition, both their low-volume production and their manufacturing complexity contribute to their elevated cost, preventing widespread adoption~\cite{robertson2018}.

\textbf{Existing light field datasets}, such as the Stanford Light Field Archives~\cite{wilburn2005high, hahne2021plenopticam, dansereau2019liff}, provide valuable resources but often lack alignment with modern plenoptic 2.0 cameras, limiting their usability. Although other datasets with similar setups exist~\cite{mousnier2015partial, rerabek2016new}, their utility is constrained by insufficient data volume, different data formats and hardware version. Hence, the lack of real-world datasets for plenoptic 2.0 cameras has led most research to rely on synthetic datasets, which, while valuable, lack the realism needed for robust deployment~\cite{oldStanford1996, mitLFarchive, honauer2016benchmark, Paudyal2016374, ShekharBZCPMKMD18, Vamsi2017}.

\textbf{Depth completion} methods generate dense depth maps from sparse representations, e.g.\ in the context of LIDAR data~\cite{khan2022comprehensive}. These methods typically refine and densify sparse depth maps using unguided techniques, such as interpolation or hand-crafted features, within the same modality~\cite{ferstl2013image, matsuo2015depth, bai2020depthnet}. However, image-guided methods have proven more successful, especially with noisy depth data~\cite{9429178, 9561035, 9918022}. %While our work involves depth estimation from plenoptic images, it yields similarities to the depth completion problem.
In the case of a light field camera, sparse depth maps are generated from a limited number of microlenses in the sensor. Integrating foundation models as an analogous image-guided method offers a promising approach for depth completion in our domain~\cite{izquierdo2023sfm}.

\begin{figure*}
    \centering
    \includegraphics[width=\textwidth]{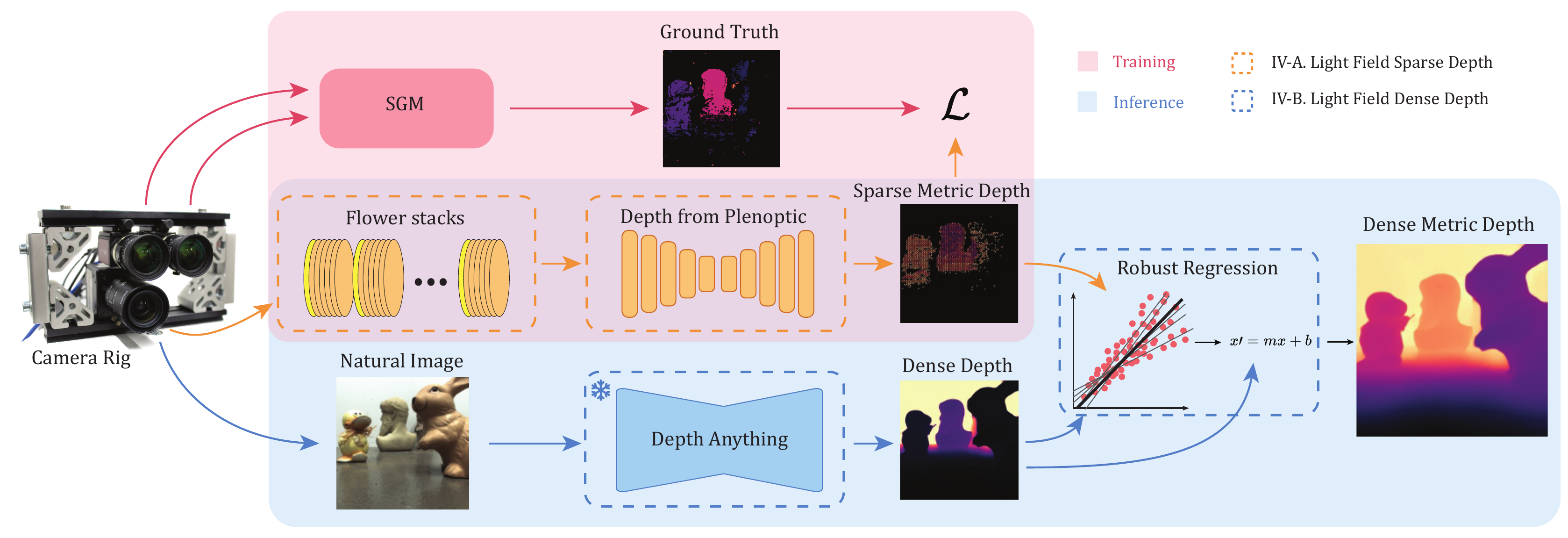}
    \caption{Overview. The Image Processing Toolkit pre-processes captured data, which is then used to train the \textit{Microlens Depth Network} for sparse metric depth estimation. These are used to refine the inference by Depth Anything, producing a dense metric depth map. Filtered stereo depth serves as the ground truth, following densification and scale alignment.}
    \label{fig:pipeline}
\end{figure*}

\section{The Light Field \& Stereo (LFS) Dataset }\label{sec:dataset}

\textbf{\emph{Dataset specifications.}} Our LFS dataset comprises images captured with a plenoptic camera and a stereo. It includes 59 captures from various static scenes in controlled laboratory conditions. Each set contains a \textit{plenoptic image}, a \textit{plenoptic virtual depth image},\footnote{Virtual depths are the measured depths of focused projections within the camera relative to the MLA plane, expressed in multiples of the (secretive) sub-millimetric distance between the MLA and the sensor chip. These virtual images resemble disparity maps in stereo vision, as they are closely tied to image processing steps while still related to actual metric depth.} and a \textit{natural image}, as well as the \textit{depth from stereo}. Plenoptic images display a pattern of 8,837 microlenses on the sensor, while natural images represent the reconstructed image from the camera's central point of view. Both plenoptic and natural images are produced using the manufacturer's proprietary software, see an example in Fig. \ref{fig:teaser}. We extract semi-dense metric depth from the calibrated stereo, that will serve as a ground-truth.

\textbf{\emph{Hardware and software.}} The cameras are mounted on a robust mechanical framework made of aluminium profiles. The setup includes two Allied Vision Mako G-419C cameras~\cite{makos} and a Raytrix R5 plenoptic camera~\cite{raytrixProducts}, which is based on the Baumer HXG40c model. The Mako cameras are compact, industrial vision cameras, while the Raytrix R5 is a light field camera with up to 1 MP effective output resolution and 4.2 Megarays light field resolution. All cameras feature a CMOS CMV4000 sensor at 4.2 MP resolution, supporting 25 fps under the GigE vision standard. The three cameras are connected to a notebook through a 2.5G switch and a 2.5G Ethernet adapter. To ensure overlapping fields of view between the stereo pair and the plenoptic camera, they are positioned close together with rigid alignment and matching lenses. Synchronization is achieved by using the plenoptic camera’s exposure signal as a master trigger for the stereo. 
The configuration, control, and data processing utilizes the vendors' SDKs, which rely on GenICam transport layer (GenTL) libraries to communicate with the cameras. This setup is extended with in-house packages for concurrent deployment of processes, frame distribution via shared memory, image post-processing, and efficient storage. Geometric camera calibration is carried out using the method by Zhang, Sturm, and Maybank~\cite{zhang:2000trpami,sturm:1999cvpr} implemented in the camera calibration toolbox DLR CalLab~\cite{callab}, along with the stepwise plenoptic camera calibration method~\cite{strobl2016stepwise} and the extrinsic calibration method by Strobl and Hirzinger~\cite{strobl2006iros}. The RxLive software from Raytrix is also utilized to capture plenoptic images and visualize scene representations, with custom methods developed for pre-processing plenoptic images. Metric stereo depth is estimated by an in-house implementation of the SGM algorithm~\cite{hirsch:2008trpami}. Lastly, custom scripts were developed for stereo image reprojection.

\textbf{\emph{Scene configuration.}} Our dataset images a variety of scenes, to benchmark generalization. It includes objects with diverse shapes and textures placed at multiple distances to avoid overfitting. The setup optimizes camera positioning and focal distances across a range of working distances. Illumination conditions are kept constant.

\begin{figure*}
    \centering
    \includegraphics[width=.95\textwidth]{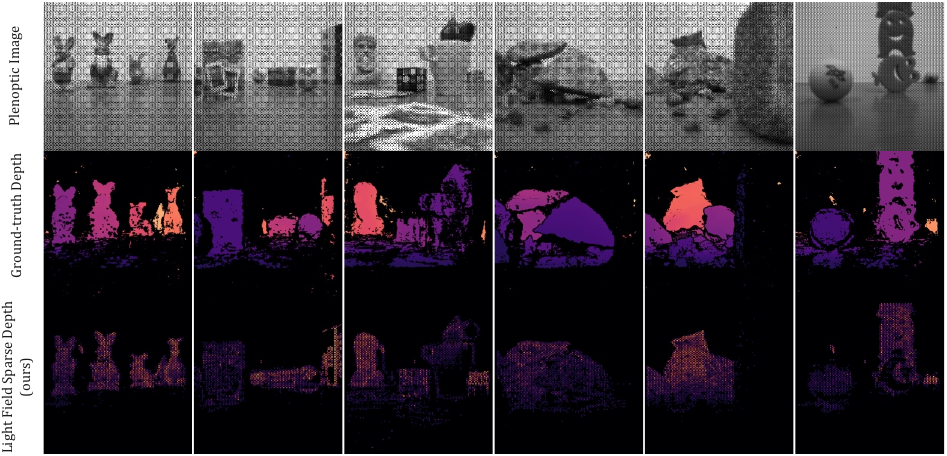}
    \caption{Qualitative sparse depth results. Top row: Plenoptic images. Center row: Ground-truth depth from stereo. Bottom row: Depth from our \textit{Microlens Depth Network}. Note how our depth values are very close to the ground truth.
    }
    \label{fig:sparse_results}
\end{figure*}

\textbf{\emph{Plenoptic images.}} We created an \textit{Image Processing Toolkit} that generates a dataset from plenoptic images by modeling the microlenses pattern for efficient indexing and manipulation. Utilizing a hexagonal grid storage system, it uses a dual coordinate system to accommodate the microlens arrangement with interlaced rows (see Fig.~\ref{fig:flower_stack}). A parametric model calculates the coordinates of each microlens based on a 2D calibration of the microlenses pattern. First, the original plenoptic image undergoes debayering into a full-color image. Other key operations include \textit{cropping}, where images are extracted based on centroid positions, and \textit{stacking}, which arranges cropped microlens images from concentric rings to create a \textit{flower stack} (tensor) for enhanced feature extraction and efficient correspondence search.

\textbf{\emph{Plenoptic depth.}} The data obtained from RxLive is processed as an alternative ground-truth depth source for benchmarking purposes. This involves decoding the virtual depth values obtained from the manufacturer and transforming them into metric ones using a thin lens camera model~\cite{strobl2016stepwise}. 

\textbf{\emph{Depth from stereo.}} The stereo processing pipeline (Fig.~\ref{fig:stereoPipeline}) involves debayering, undistortion, rectification, and the SGM algorithm (correspondence search, triangulation, and regularization) to produce disparity and metric depth maps. The resulting RGBD image is then re-projected, as a colored pointcloud, onto the calibrated light field camera pose, synthesizing distorted color and metric depth images. While the resulting images resemble natural images, discontinuities occur due to occlusions and untextured regions. To reduce noise in low-texture areas, regularization measures like consistency checks, gradient filters, and hierarchical matching are applied. Finally, depth values are extracted at microlens centroid coordinates of the respective plenoptic image using the \textit{Image Processing Toolkit}.

\begin{table*}[t]
    \caption{Comparison of our Light Field Sparse and Dense depth against 1) A Random Depth generator, 2) the Sparse Depth provided by the Raytrix manufacturer software, and 3) the foundation model Depth Anything. %Note how both our sparse and dense modules outperform the chosen baselines.
    }
    \centering
    \resizebox{0.8\textwidth}{!}{%
    \begin{tabular}{|l|c|c|c|c|c|c|c|c|} 
    \hline
                        & \textbf{MSE [cm$^2$] $\downarrow$} & \textbf{RMSE [cm] $\downarrow$} & \textbf{MARE $\downarrow$} & \textbf{MSRE $\downarrow$} & \textbf{$\delta$ $\uparrow$} & \textbf{$\delta^2$ $\uparrow$} & \textbf{$\delta^3$ $\uparrow$} & \textbf{BPR $\downarrow$}  \\ 
    \hline
    Random Depth        & 1551.39                   & 39.27                      & 1403.74                    & 23.45                      & -                            & -                              & -                              & -                          \\ 
    \hline\hline
    Light Field Sparse Depth (Raytrix)~\cite{rxlive}       & 136.47                    & 10.94                      & 65.43                      & 2.70                       & \textbf{88.62}               & \textbf{93.04}                 & \textbf{95.43}                 & \textbf{0.3043}             \\ 
    \hline
    Light Field Sparse Depth (our \textit{Microlens Depth Network})       & \textbf{124.68}           & \textbf{5.55}              & \textbf{52.75}             & \textbf{2.63}              & 84.83                        & 88.40                          & 91.25                          & 0.3949                     \\ 
    \hline\hline
    Depth Anything~\cite{depthanything}      & 129.44                    & 10.96                      & 40.90                      & 5.24                       & 18.30                        & 38.70                          & 65.00                          & 0.8623                     \\ 
    \hline
    Light Field Dense Depth (ours)          & \textbf{83.21}            & \textbf{8.50}              & \textbf{37.30}             & \textbf{3.94}              & \textbf{46.40}               & \textbf{74.60}                 & \textbf{90.00}                 & \textbf{0.4233}             \\ 
    \hline
    \end{tabular}%
    }
    \label{tab:final_results}
\end{table*}

\section{Single-Shot Depth from Plenoptic Cameras}

We present a complete pipeline to obtain dense, metric depth maps from a single light field image, %The optical configuration of the plenoptic camera is leveraged to treat the microlens array as a multi-view stereo, providing the disparity to estimate metric depth. The proposed pipeline 
that combines image processing, microlens-scale depth estimation (Section \ref{sec:sparse-depth}), and subsequent densification and refinement to generate a dense metric depth map (Section \ref{sec:scale}), see Fig.~\ref{fig:pipeline}.

\subsection{Sparse Depth from Plenoptic Images}\label{sec:sparse-depth}

Our model follows an encoder-decoder convolutional architecture. \textit{Flower stacks} are provided to the encoder in the form of 4D tensors $X_i$ of size $(N, C, H, W)$, for which $N$ is the batch size, $C$ the number of channels, and $H \!\times\! W$ the height and width of each microlens projection onto the sensor. Our model infers a single metric depth value per stack, corresponding to the prediction at the centroid of the main microlens of each stack.
This network, which we denote as \textit{Microlens Depth Network}, is composed of 2D convolutions, batch normalization, activation functions, and fully-connected bottleneck layers. It captures spatial relationships among microlenses, leveraging their redundancy in the \textit{flower stack} to enhance depth estimation robustness, e.g.\ in regions with occlusions.
The estimated depths form a sparse depth map aligned with the central sub-aperture image at a lower resolution due to the limited number of microlenses (8,837).
The resulting depth map is inaccurate in regions with weak textures, which we then remove by applying a texture filter.

\subsection{Dense Depth from Plenoptic Cameras}\label{sec:scale}

In order to obtain dense per-pixel metric depth values, we fuse these results with the pretrained model \textit{Depth Anything}~\cite{depthanything}, which generates dense disparity maps from monocular images. \textit{Depth Anything} is %based on the Vision Transformer (ViT) architecture and 
trained on extensive datasets, ensuring robustness across a variety of scenarios, and showing state-of-the-art performance in the most recent benchmarks~\cite{spencer2024third}. However, its disparity maps are not in metric scale. We use a robust regression method to align them to the sparse metric depth values predicted by our \textit{Microlens Depth Network}. We use the \textit{Theil-Sen} estimator~\cite{theil1950rank,sen1968} for this alignment, as it handles outliers in an effective manner.
First, dense disparity values are extracted at locations with corresponding sparse metric depths (the latter are converted to disparities for consistency, to achieve a monotonic function that preserves the order), and then aligned using the \textit{Theil-Sen} approach. This non-parametric method calculates the median slope \( m \) from all pairs of points \( (x_i, y_i) \) and \( (x_j, y_j) \) as follows:
\begin{equation}
    m = \mathrm{median} \left\{ \frac{y_j - y_i}{x_j - x_i} \mid x_i \neq x_j \right\} \quad .
\end{equation}
The intercept term \( b \) is then determined by taking the median of the individual intercepts
\begin{equation}
    b = \mathrm{median} \{ y_i - m x_i \} \quad ,
\end{equation}
resulting in the linear model $y = m x + b$.
%
%\begin{equation}
%    y = m x + b \quad .
%\end{equation}
%
This linear model is then applied to scale and offset the depths predicted by Depth Anything. Finally, disparities are transformed back to metric depth using the intrinsics of the rectified stereo camera.

\subsection{Implementation Details}

\textbf{\emph{Architecture.}} The encoder of the \textit{Microlens Depth Network} comprises five convolutional layers with 2D convolutions, batch normalization, and ReLU activations, processing input tensors of shape \(X_i = (N, C, H, W)\), where the batch size is $N \!=\! 128$. The \textit{flower stacks} are of size $23 \!\times\! 23$ pixels with seven RGB images each, resulting in $C \!=\! 21$ channels. The bottleneck includes a multilayer perceptron (MLP) with three fully connected layers, compressing the encoded data into a lower-dimensional space to capture high-level features like correspondences and disparities. The decoder, which mirrors the encoder, uses five transposed convolutional layers, with the final layer outputting a single depth channel.

\textbf{\emph{Loss.}} We use the mean squared error (MSE) between our per-pixel depth predictions \(\hat{y}\) and their ground-truth values \(y\) 
\begin{equation}
    \mathit{MSE} = \frac{1}{\sum_{i=1}^{n} M_i} \sum_{i=1}^{n} M_i (\hat{y}_i - y_i)^2 \quad ,
\end{equation}
where \(M_i\) is a binary mask indicating whether the ground-truth value at pixel \(i\) is available (1) or not (0).

\begin{figure*}
    \centering
    \includegraphics[width=.9\textwidth]{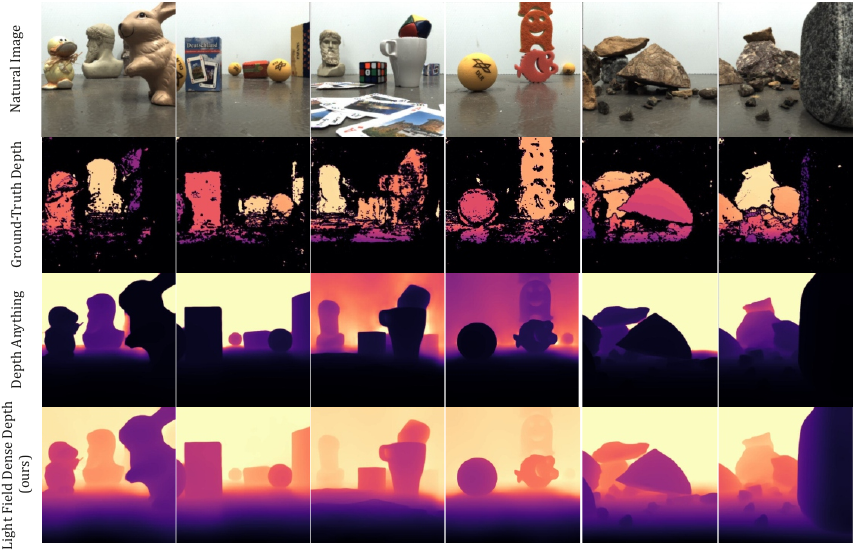}
    \caption{Qualitative results. From natural images \textit{(first row)}, Depth Anything infers dense depth (yet up-to-scale) \textit{(third row)}. Then, scale alignment with metric values from our Sparse Light Field Depth transforms it into a dense metric depth \textit{(fourth row)}. Note our Light Field Dense depth aligns more closely with the ground truth depth \textit{(second row)} than Depth Anything.}
    \label{fig:dense_results}
\end{figure*}

\section{Experimental Results}

\subsection{Data Setup} 

As detailed in Section \ref{sec:dataset}, the LFS dataset consists of 59 images captured, each containing 8,837 microlenses, resulting in 8,465 \textit{flower stacks}\footnote{The generation of \textit{flower stacks} is only carried out if all microlenses are within the image boundaries. Otherwise, the stack is discarded.} of 7 RGB crops, with a resolution of $23 \!\times\! 23$ pixels. 
We split the dataset image-wise in train and test with 49 and 10 images respectively. In Tables \ref{tab:final_results} and \ref{tab:ablation} we report the standard metrics used by the single-view depth community, find their definitions here~\cite{eigen2014depth}.

\subsection{Training Details} 
We train the \textit{Microlens Depth Network} for 125 epochs using batches of 128 \textit{flower stacks}. This requires around 10 hours using a Quadro FV100 Volta GPU. We optimize using Adam \cite{kingma2014adam} with a learning rate of $0.001$. % and apply early stopping with 20 epochs patience.
For \textit{Depth Anything}, we used $1024 \!\times\! 1024$ RGB images. \textit{Theil-Sen} regression was applied to align scale and offset using 11,002 valid depth values out of 17,321 sparse data points. The process took 1.78 seconds per image on the same GPU.

\subsection{Comparison against Baselines}

Our \textit{Microlens Depth Network} took 10.54 minutes for a test set of 84,650 \textit{flower stacks}, representing 20\% of the LFS dataset (10 images). The test images encompass diverse scenes with elements at varying distances. Table~\ref{tab:final_results} shows the aggregated results, along with those of related baselines.

We first report the results of a dummy baseline using randomly generated depth values, to properly assess the results of the methods. After that, note how our sparse depth achieves a RMSE of 5.55 cm, almost halving the error over the software provided by the camera manufacturer (Raytrix) with an additional conversion from virtual to metric depths~\cite{rxlive,strobl2016stepwise}. Finally, observe how our metrics significantly improve the ones of \textit{Depth Anything}, the state of the art on single-view depth estimation, as this last model cannot observe the metric scale of the scene. The accuracy in the prediction of the scale, compared to that of \textit{Depth Anything}, can be clearly observed in the qualitative results of Fig.~\ref{fig:dense_results}. The source of this scale accuracy traces back to the accurately scaled predictions of our \textit{Microlens Depth Network}, which can also be qualitatively assessed in Fig.~\ref{fig:sparse_results}.

\vspace{-0.05 cm}
\begin{table*}
    \caption{Error metrics for additional methods explored. The first and fourth rows present the proposed methods. Each of these is then ablated and compared quantitatively. The proposed ones outperform the alternatives.}
    
    \centering
    \resizebox{0.748\textwidth}{!}{%
    \begin{tabular}{|l|c|c|c|c|c|c|c|c|} 
    \hline
                      & \textbf{MSE} [cm$^2$] $\downarrow$ & \textbf{RMSE [cm] $\downarrow$} & \textbf{MARE $\downarrow$} & \textbf{MSRE $\downarrow$} & \textbf{$\delta$ $\uparrow$} & \textbf{$\delta^2$ $\uparrow$} & \textbf{$\delta^3$ $\uparrow$} & \textbf{BPR $\downarrow$}  \\ 
    \hline
    Light Field Sparse Depth (ours)       & \textbf{124.68}           & \textbf{5.55}              & \textbf{52.75}             & \textbf{2.63}              & \textbf{84.83}               & 88.40                          & 91.25                          & 0.3949                     \\ 
    \hline
    Parallel encoders     & 250.18                    & 9.45                       & 104.57                     & 5.13                       & 81.02                         & \textbf{89.74}                 & \textbf{93.03}                 & 0.6872                     \\ 
    \hline
    Weighted Mask     & 171.57                    & 12.30                      & 80.82                      & 3.64                       & 83.43                         & 88.32                          & 93.01                          & \textbf{0.3802}            \\ 
    \hline\hline
    Light Field Sparse Depth (ours)        & \textbf{83.21}            & \textbf{8.50}              & 37.30                      & 3.94                       & \textbf{46.40}               & \textbf{74.60}                 & \textbf{90.00}                 & \textbf{0.4233}            \\ 
    \hline
    Huber             & 110.40                    & 9.92                       & 41.30                      & \textbf{2.96}              & 38.00                         & 73.10                          & 83.60                          & 0.4758                     \\ 
    \hline
    SGD-Huber         & 110.75                    & 9.94                       & \textbf{27.60}             & 3.12                       & 43.40                         & 73.60                          & 87.90                          & 0.4631                     \\
    \hline
    \end{tabular}%
    }
    \label{tab:ablation}
\end{table*}

\subsection{Ablation Study}

To validate our design choices, we carried out ablative experiments that assessed the importance of each. We summarized them below, and results are shown in Table~\ref{tab:ablation}.

\textbf{\emph{Flower stacks.}} In addition to our \textit{flower stacks}, we experimented with single microlens images and double-ringed \textit{flower stacks}. The former lacked multi-view context and thus produced much worse depth estimates, while the latter did not improve the accuracy significantly and had a higher computational cost.

\textbf{\emph{Network's architecture.}} We tested a fully-convolutional architecture with parallel encoders and a bottleneck, which underperformed compared to our single encoder architecture, which better captured the global context within the \textit{flower stacks}. Integrating an MLP in the bottleneck improved the model’s ability to learn complex patterns. %Lastly, removing the decoder proved detrimental, reinforcing the necessity of a decoding stage.

\textbf{\emph{Alternative ground-truth.}} Comparisons between ground truth depth from the light field camera and the stereo cameras revealed that the stereo depth, despite requiring additional pre-processing from our side, yielded better performance due to a more accurate metric calibration.

\textbf{\emph{Metric scaling. }} We adapted the test-time refinement by Izquierdo and Civera~\cite{izquierdo2023sfm}, to aggregate our sparse depth with Depth Anything. 
However, the method struggled with the multi-modal noise distribution of stereo data, which led us to test other regression techniques for robust scale alignment, including the Huber Regressor and the Stochastic Gradient Descent (SGD) with Huber loss. The results showed that Theil-Sen and RANSAC remain most effective for the specific data distributions in this problem.

\section{Conclusion}

This work introduces a novel approach for single-shot depth from a plenoptic camera 2.0. We developed a novel \textit{Microlens Depth Network} that successfully infers sparse metric depth from plenoptic images, we then synthesize a natural image from the plenoptic one, that we forward pass through the foundation single-view up-to-scale dense depth model Depth Anything. Finally, we regress the true scale and offset values to correct the dense output from our sparse depth. Our end-to-end pipeline demonstrates the feasibility of generating dense metric depth maps from single shots. Our work makes several key contributions, including the design of a specialized neural network for depth estimation, the creation and release of the Light Field \& Stereo Image Dataset (LFS) -- a new dataset with plenoptic images and corresponding reprojected metric depth labels -- and the development of a comprehensive image pre-processing methodology suited for learning-based applications. These contributions advance the state of the art in light field imaging and single-view depth estimation, establishing a foundation for further research. Future work could focus on refining the developed methodology by exploring more advanced regression techniques, that incorporate object segmentation and independent scale estimation for different image regions, alongside occlusion handling. This would enable more accurate depth scaling, improve metric precision, and address the complexities in the relationships among objects, as opposed to our linear regression approach.

\balance
{
\bibliographystyle{ieeetr}
\bibliography{paper}
}

\end{document}